\documentclass[runningheads]{llncs}
\usepackage{graphicx}
\usepackage{amsmath,amssymb}
\usepackage{color}
\usepackage{times}
\usepackage{epstopdf}
\usepackage[export]{adjustbox}
\usepackage{epsfig}

\usepackage{microtype}
\usepackage{algorithm}
\usepackage{algpseudocode}
\usepackage{multirow}
\usepackage[list=true]{subcaption}
\usepackage{caption}
\usepackage{array}
\usepackage[outercaption]{sidecap}    
\newcolumntype{C}[1]{>{\centering\let\newline\\\arraybackslash\hspace{0pt}}m{#1}}
\usepackage[width=122mm,left=12mm,paperwidth=146mm,height=193mm,top=12mm,paperheight=217mm]{geometry}
\usepackage{hyperref}
\hypersetup{colorlinks=true, linkcolor=blue, filecolor=blue, urlcolor=blue}
\begin{document}

\title{W-TALC: Weakly-supervised Temporal Activity Localization and Classification} 

\titlerunning{W-TALC: Weakly-supervised Temporal Activity Localization and Classification}

%
\author{Sujoy Paul, Sourya Roy \and Amit K Roy-Chowdhury}
%
\authorrunning{S. Paul, S. Roy and A. K. Roy-Chowdhury}
%

\institute{University of California, Riverside, CA 92521, USA
\email{\{supaul,sroy,amitrc\}@ece.ucr.edu}}

\maketitle              

{\let\thefootnote\relax\footnote{{European Conference on Computer Vision (ECCV), 2018}}}
\vspace*{-10mm}

\begin{abstract}
Most activity localization methods in the literature suffer from the burden of frame-wise annotation requirement. Learning from weak labels may be a potential solution towards reducing such manual labeling effort. Recent years have witnessed a substantial influx of tagged videos on the Internet, which can serve as a rich source of weakly-supervised training data. Specifically, the correlations between videos with similar tags can be utilized to temporally localize the activities. Towards this goal, we present W-TALC, a Weakly-supervised Temporal Activity Localization and Classification framework using only video-level labels. The proposed network can be divided into two sub-networks, namely the Two-Stream based feature extractor network and a weakly-supervised module, which we learn by optimizing two complimentary loss functions. Qualitative and quantitative results on two challenging datasets - Thumos14 and ActivityNet1.2, demonstrate that the proposed method is able to detect activities at a fine granularity and achieve better performance than current state-of-the-art methods. Codes available at \href{https://github.com/sujoyp/wtalc-pytorch}{https://github.com/sujoyp/wtalc-pytorch}

\end{abstract}
%
%
%

\section{Introduction}
\label{intro}
\begin{SCfigure}
	\centering
	\includegraphics[height=4.1cm,width=5.9cm]{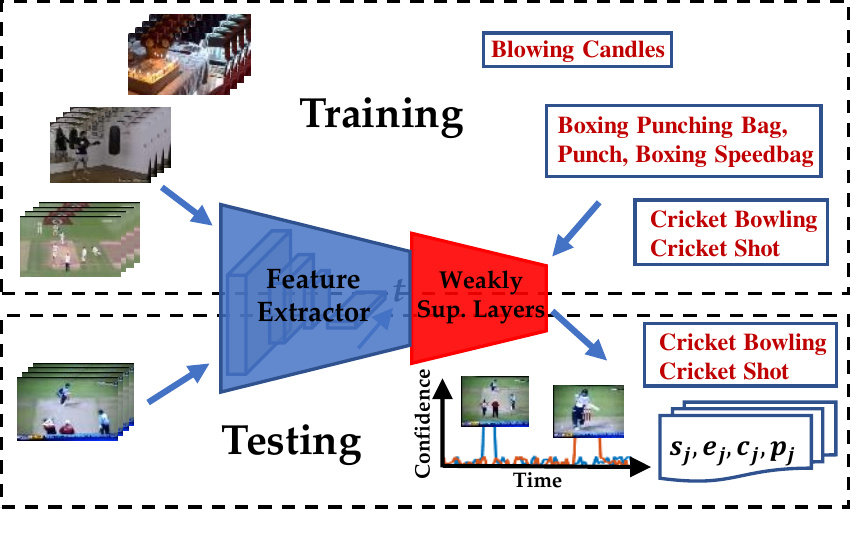}	
	\caption{This figure presents the train-test protocol of W-TALC. The training set consists of videos and the corresponding video-level activity tags. Whereas, while testing, the network not only estimates the labels of the activities in the video, but also temporally locates their occurrence representing the start ($s_j$) and end time ($e_j$), category ($c_j$) and confidence of recognition ($p_j$) of the $j^{th}$ activity located by the model.}
	\label{motivation}
\end{SCfigure}

Temporal activity localization and classification in continuous videos is a challenging and interesting problem in computer vision \cite{aggarwal2011human}. Its recent success \cite{xu2017r,zhao2017temporal} has evolved around a \textit{fully} supervised setting, which considers the availability of frame-wise activity labels. However, acquiring such precise frame-wise information requires enormous manual labor. This may not scale efficiently with a growing set of cameras and activity categories. On the other hand, it is much easier for a person to provide a few labels which encapsulate the content of a video. Moreover, videos available on the Internet are often accompanied by tags which provide semantic discrimination. Such video-level labels are generally termed as \textit{weak} labels, which may be utilized to learn models with the ability to classify and localize activities in continuous videos. In this paper, we propose a novel framework for Temporal Activity Localization and Classification (TALC) from such weak labels. Fig. \ref{motivation} presents the train-test protocol W-TALC. 

In computer vision, researchers have utilized weak labels to learn models for several tasks including semantic segmentation \cite{hartmann2012weakly,khoreva2017simple,yan2017weakly}, visual tracking \cite{zhong2014visual}, reconstruction \cite{tulyakov2017weakly,kanazawa2016warpnet}, video summarization \cite{panda2017weakly}, learning robotic manipulations \cite{singh2017gplac}, video captioning \cite{shen2017weakly}, object boundaries \cite{khoreva2016weakly}, place recognition \cite{arandjelovic2016netvlad}, and so on. The weak TALC problem is analogous to weak object detection in images, where object category labels are provided at the image-level. There have been several works in this domain mostly utilizing the techniques of Multiple Instance Learning (MIL) \cite{zhou2004multi} due to their close relation in terms of the structure of information available for training. 
The positive and negative bags required for MIL are generated by state-of-the-art region proposal techniques \cite{li2016weakly,jie2017deep}. On the other hand, end-to-end learning with categorical loss functions are presented in \cite{durand2017wildcat,durand2016weldon,diba2016weakly,singh2017hide} and recently, the authors in \cite{zhu2017soft} incorporated the proposal generation network in an end-to-end manner. 

Temporal localization using weak labels is a much more 
challenging task compared to weak object detection. The key reason is the additional variation in content as well as the length along the temporal axis in videos. 
Activity localization from weakly labeled data remains relatively unexplored. Some works \cite{siva2011weakly,yan2017weakly,sultani2016if} focus on weakly-supervised spatial segmentation of the actor region in short videos. Another set of works \cite{bojanowski2014weakly,kuehne2017weakly,richard2017weakly,huang2016connectionist} considers video-level labels of the activities and their temporal ordering during training. However, such information about the activity order may not be available for a majority of web-videos. A recent work \cite{weinzaepfel2017human} utilizes state-of-the-art object detectors for spatial annotations but considers full temporal supervision. In \cite{wang2017untrimmednets}, a soft selection module is introduced for untrimmed video classification along with activity localization and a sparsity constraint is included in \cite{nguyen2017weakly}.

In W-TALC, as we have labels only for the entire video, we need to process them at once. Processing long videos at fine temporal granularity may have considerable memory and computation requirements. On the other hand, coarse temporal processing may result in reduced detection granularity. Thus, there is a trade-off between performance and computation. Over the past few years, networks trained on ImageNet \cite{deng2009imagenet} and recently on Kinetics \cite{kay2017kinetics}, has been used widely in several applications. Based on these advances in literature and the aforementioned trade-off, we may want to ask the question that: \textit{is it possible to utilize these networks just as feature extractors and develop a framework for weakly-supervised activity localization which learns only the task-specific parameters, thus scaling up to long videos and processing them at fine temporal granularity?} To address this question, in this paper, \textit{we present a framework (W-TALC) for weakly-supervised temporal activity localization and video classification, which utilizes pair-wise video similarity constraints via an attention-based mechanism along with multiple instance learning to learn only the task-specific parameters.}

\begin{figure*}[t]
	\centering
	\includegraphics[scale=0.395]{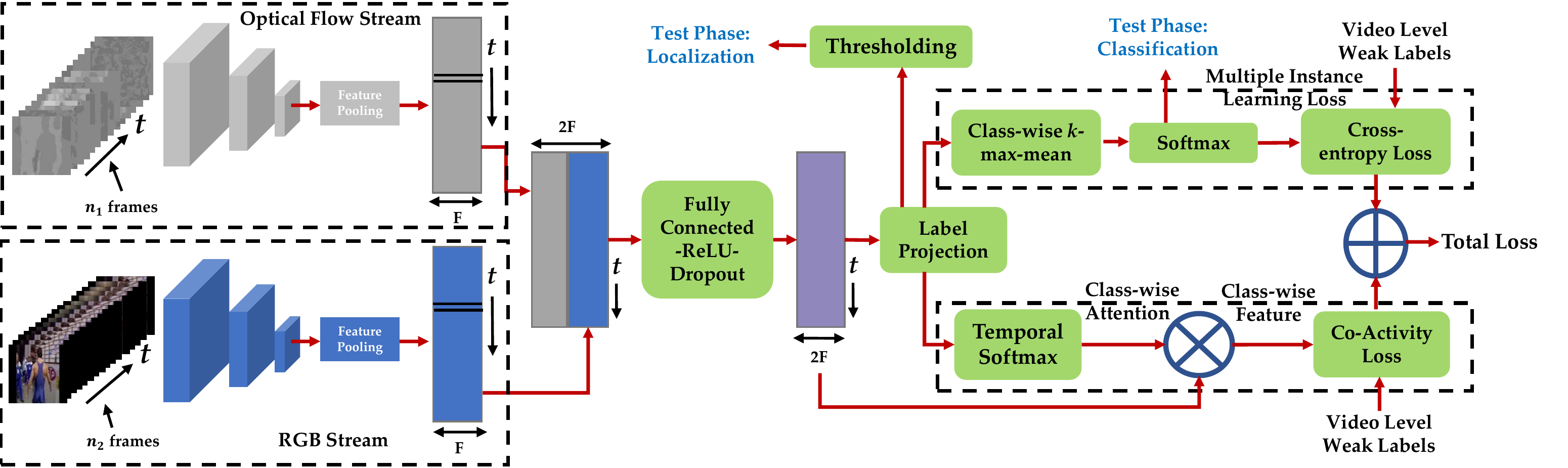}
	\caption{This figure presents the proposed framework for weakly-supervised activity localization and classification. The number of frames $n_1$ and $n_2$ are dependent on the feature extractor used. After concatenating the feature vectors from the RGB and Optical Flow streams, a FullyConnected-ReLU-Dropout operation is applied to get features of dimension $2048$ for each time instant. These are then passed through the label projection module to obtain activations over the categories.  Using these activations, we compute two loss functions namely Multiple Instance Learning Loss and Co-Activity Similarity Loss, which are optimized jointly to learn the network weights.}
	\label{framework}
\end{figure*}

\textbf{Framework Overview.}
A pictorial representation of W-TALC is presented in Fig. \ref{framework}. The proposed method utilizes off-the-shelf Two-Stream networks (\cite{wang2017untrimmednets,carreira2017quo}) as a feature extractor. The number of frame inputs depend on the network used and will be discussed in Section \ref{feature}. 
After passing the frames through the networks, we obtain a matrix of feature vectors with one dimension representing the temporal axis. Thereafter, we apply a FullyConnected-ReLU-Dropout layer followed by label space projection layer, both of which is learned for the weakly-supervised task. 

The activations over the label space are then used to compute two complimentary loss functions using video-level labels. The first one is Multiple Instance Learning Loss, where the class-wise $k$-max-mean strategy is employed to pool the class-wise activations and obtain a probability mass function over the categories. Its cross-entropy with the ground-truth label is the Multiple Instance Learning Loss (MILL). The second one is the Co-Activity Similarity Loss (CASL), which is based on the motivation that a pair of videos having at least one activity category (say biking) in common should have similar features in the temporal regions which correspond to that activity. Also, the features from one video corresponding to biking should be different from features of the other video (of the pair) not corresponding to biking. However, as the temporal labels are not known in weakly-supervised data, we use the attention obtained from the label space activations as weak temporal labels, to compute the CASL. Thereafter, we jointly minimize the two loss functions to learn the network parameters. 

\textbf{Main contributions.} The main contributions of the proposed method are as follows. 1. We propose a novel approach for weakly-supervised temporal activity localization and video classification, without fine-tuning the feature extractor, but learning only the task-specific parameters. Our method does not consider any ordering of the labels in the video during training and can detect multiple activities in the same temporal duration.\\
2. We introduce the Co-Activity Similarity Loss and jointly optimize it with the Multiple Instance Learning Loss to learn the network weights specific to the weakly-supervised task. We empirically show that the two loss functions are complimentary in nature. \\
3. We perform extensive experimentations on two challenging datasets and show that the proposed method performs better than the current state-of-the-art methods. 

\section{Related Works.}
The problem of learning from weakly-supervised data has been addressed in several computer vision tasks including object detection \cite{bilen2016weakly,durand2016weldon,li2016weakly,shi2017transfer,cinbis2017weakly,singh2017hide}, segmentation \cite{vezhnevets2010towards,pathak2015constrained,bearman2016s,khoreva2017simple,wei2017stc}, video captioning \cite{shen2017weakly} and summarization \cite{panda2017weakly}. Here, we discuss in detail the other works which are more closely related to our work.

\textbf{Weakly-supervised Spatial Action Localization.} Some researchers have looked into the problem of spatial localization of actors in mostly short and trimmed videos using weak supervision. In \cite{chen2017attending} a framework is developed for localization of players in sports videos, using detections from state-of-the-art fully supervised player detector, as inputs to their network. Person detectors are also used in \cite{siva2011weakly,weinzaepfel2016towards} to generate person tubes, which is used to learn different Multiple Instance Learning based classifiers. Conditional Random Field (CRF) is used in \cite{yan2017weakly} to perform actor-action segmentation from video-level labels but on short videos. 

\textbf{Scripts as Weak Supervision.} Some works in the literature use scripts or subtitles generally available with videos as weak labels for activity localization. In \cite{laptev2008learning,duchenne2009automatic} words related to human actions are extracted from subtitles to provide coarse temporal localizations of actions for training. In \cite{bojanowski2013finding}, actor-action pairs extracted from movie scripts serve as weak labels for spatial actor-action localization by using discriminative clustering. Our algorithm on the other hand only considers that the label of the video is available as a whole, agnostic to the source from where the labels are acquired, i.e., movie scripts, subtitles, humans or other oracles.

\textbf{Temporal Localization with Ordering.} Few works in the literature have considered the availability of temporal order of activities, apart from the video-level labels during training. The activity orderings in the training videos are used as constraints in discriminative clustering to learn activity detection models in \cite{bojanowski2014weakly}. A similar approach was taken in \cite{bojanowski2015weakly}. In \cite{huang2016connectionist}, the authors propose a dynamic programming based approach to evaluate and search for possible alignments between video frames and the corresponding labels. The authors in \cite{richard2017weakly} use a Recurrent Neural Network (RNN) to iteratively train and realign the activity regions until convergence. A similar iterative process is presented by the same authors in \cite{kuehne2017weakly}, but without employing an RNN. Unlike these works in literature, our work does not consider any information about the orderings of the activity.

The works in \cite{wang2017untrimmednets,nguyen2017weakly} are closely related to the problem setting presented in this paper. However, as the framework in \cite{wang2017untrimmednets} is based on the temporal segments network \cite{wang2016temporal}, a fixed number of segments, irrespective of the length of the video, are considered during training, which may lead to a reduction in localization granularity. Moreover, they only employ the MILL, which may not be enough to localize activities at fine temporal granularity. A sparsity-based loss function is optimized in \cite{nguyen2017weakly}, along with a loss function similar to that obtained using the soft selection method in \cite{wang2017untrimmednets}. In this paper, we introduce a novel loss function named Co-Activity Similarity Loss (CASL) which imposes pair-wise constraints for better localization performance. We also propose a mechanism for dealing with long videos and yet detecting activities at high temporal granularity. In spite of not finetuning the feature extractor, we can still achieve better performance than state-of-the-art methods on weak TALC. Moreover, experimental results show that CASL is complimentary in nature with MILL. 

\section{Methodology}
In this section, we present our framework (W-TALC) for weakly-supervised activity localization and classification. First, we present the mechanism we use to extract features from the two standard networks, followed by the layers of the network we learn. Thereafter, we present two loss functions MILL and CASL, which we jointly optimize to learn the weights of the network. It may be noted that we compute both the loss functions using only the video-level labels of training videos. Before going into the details of our framework, let us define the notations and problem statement formally. 

\textbf{Problem Statement.} Consider that we have a training set of $n$ videos $\mathcal{X}=\{\boldsymbol{x}_i\}_{i=1}^n$ with variable temporal durations denoted by $\boldsymbol{L}=\{l_i\}_{i=1}^n$ (after feature extraction) and activity label set $\boldsymbol{A}=\{\boldsymbol{a}_i\}_{i=1}^n$, where $\boldsymbol{a}_i=\{a_i^j\}_{j=1}^{m_i}$ are the $m_i (\geq 1)$ labels for the $i^{th}$ video. We also define the set of activity categories as $\mathcal{S}=\bigcup_{i=1}^n\boldsymbol{a}_i=\{\alpha_i\}_{i=1}^{n_c}$. During test time, given a video $\boldsymbol{x}$, we need to predict a set $\boldsymbol{x}_{det}=\{(s_j,e_j,c_j,p_j)\}_{j=1}^{n(\boldsymbol{x})}$, where $n(\boldsymbol{x})$ is the number of detections for $\boldsymbol{x}$. $s_j,e_j$ are the start time and end time of the $j^{th}$ detection, $c_j$ represents its predicted activity category with confidence $p_j$. With these notations, our proposed framework is presented next. 

\subsection{Feature Extraction}
\label{feature}
In this paper, we focus particularly on two architectures - UntrimmedNets \cite{wang2017untrimmednets} and I3D \cite{carreira2017quo} for feature extraction, mainly due to their two stream nature, which incorporates rich temporal temporal information in one of the streams, necessary for activity recognition. Please note that the rest of our framework is agnostic to the features used.

\textbf{UntrimmedNet Features.} In this case, we pass one frame through the RGB stream and 5 frames through the Optical Flow stream as in \cite{wang2017untrimmednets}. We extract the features from just before the classification layer at 2.5 fps. We use the network which is pre-trained on ImageNet \cite{deng2009imagenet}, and finetuned using weak labels and MILL for task-specific dataset as in \cite{wang2017untrimmednets}. Thus, this feature extractor has no knowledge about activities using strong labels. 

\textbf{I3D Features.} As in \cite{nguyen2017weakly}, we also experiment with features extracted from the Kinetics pre-trained I3D network \cite{carreira2017quo}. The input to the two streams are non-overlapping $16$ frame chunks. The output is passed through a 3D average pooling layer of kernel size $2 \times 7 \times 7$ to obtain features of dimension $1024$ each from the two streams. 

At the end of the feature extraction procedure, each video $\boldsymbol{x}_i$ is represented by two matrices $\boldsymbol{X}^r_i$ and $\boldsymbol{X}^o_i$, denoting the RGB and optical flow features respectively, both of which are of dimension $1024 \times l_i$. Note that $l_i$ is not only dependent on the video index $i$, but also on the feature extraction procedure used. These matrices become the input to our weakly-supervised learning module.

\textbf{Memory Constraints.} As mentioned previously, natural videos may have large variations in length, from a few seconds to more than an hour. In the weakly-supervised setting, we have information about the labels for the video as a whole, thus requiring it to process the entire video at once. This may be problematic for very long videos due to GPU memory constraints. A possible solution to this problem may be to divide the videos into chunks along the temporal axis \cite{wang2016temporal} and apply a temporal pooling technique to reduce the length of each chunk to a single representation vector. The number of chunks depends on the available GPU memory. However, this will introduce unwanted background activity feature in the representation vectors as the start and end period of the activities in the video will not overlap with the pre-defined chunks for most of the videos. To cope with this problem, we introduce a simple video sampling technique.

\textbf{Long Video Sampling.} As granularity of localizations is important for activity localization, we take an approach alternative to the one mentioned above. We process the entire video if its length is less than the pre-defined length $T$ necessary to meet the GPU bandwidth. However, if the length of the video is greater than $T$, we randomly extract from it a clip of length $T$ with contiguous frames and assign all the labels of the entire video to the extracted video clip. It may be noted that although this may introduce some errors in the labels, this way of sampling does have advantages, as will be discussed in more detail in Section \ref{exp}. 

\textbf{Computational Budget and Finetuning.} The error introduced by the video sampling strategy will increase with a decrease in the pre-defined length $T$, which meet the GPU bandwidth constraints. If we want to jointly finetune the feature extractor along with training our weakly-supervised module, $T$ may be very small in order to maintain a reasonable batch size for Stochastic Gradient Descent (SGD) \cite{bottou2010large}. Although the value of $T$ may be increased by using multiple GPUs simultaneously, it may not be a scalable approach. Moreover, the time to train both the modules may be high. Considering these problems, we do not finetune the feature extractors, but only learn the task-specific parameters, described next, from scratch. The advantages for doing this are twofold - the weakly-supervised module is light-weight in terms of the number of parameters, thus requiring less time to train, and it increases $T$ considerably, thus reducing labeling error while sampling long videos. We next present our weakly-supervised module.

\subsection{Weakly Supervised Layer}

In this section, we present the proposed weakly-supervised learning scheme, which uses only weak labels to learn models for simultaneous activity localization and classification. 

\textbf{Fully Connected Layer.} We introduce a fully connected layer followed by ReLU \cite{nair2010rectified} and Dropout \cite{srivastava2014dropout} on the extracted features. The operation can be formalized for a video with index $i$ as follows.

\begin{equation}
\boldsymbol{X}_i = \mathcal{D}\Big( \max\Big(0,\boldsymbol{W}_{fc}
\begin{bmatrix}
\boldsymbol{X}_i^r \\
\boldsymbol{X}_i^o \\
\end{bmatrix} \oplus \boldsymbol{b}_{fc}\Big),k_p
\Big)
\label{featrep}
\end{equation}
where $\mathcal{D}$ represents \texttt{Dropout} with $k_p$ representing its keep probability, $\oplus$ is the addition with broadcasting operator,  $\boldsymbol{W}_{fc} \in \mathbb{R} ^{2048 \times 2048}$ and $\boldsymbol{b} \in \mathbb{R}^{2048 \times 1}$ are the parameters to be learned from the training data and $\boldsymbol{X}_i \in \mathbb{R}^{2048 \times l_i}$ is the output feature matrix for the entire video.

\textbf{Label Space Projection}
We use the feature representation $\boldsymbol{X}_i$ to classify and localize the activities in the videos. We project the representations $\boldsymbol{X}_i$ to the label space ($\in \mathbb{R}^{n_c}, \ n_c$ is the number of categories), using a fully connected layer, with weight sharing along the temporal axis. The class-wise activations we obtain after this projection can be represented as follows.

\begin{equation}
\boldsymbol{\mathcal{A}}_i = \boldsymbol{W}_{a}\boldsymbol{X}_i \oplus \boldsymbol{b}_a
\label{act}
\end{equation}
where $\boldsymbol{W}_a \in \mathbb{R}^{n_c \times 2048}$, $\boldsymbol{b}_a \in \mathbb{R}^{n_c}$ are to be learned and $\boldsymbol{\mathcal{A}}_i \in \mathbb{R}^{n_c \times l_i}$. These class-wise activations represent the possibility of activities at each of the temporal instants. These activations are used to compute the loss functions as presented next. 

\subsection{$k$-max Multiple Instance Learning}
As discussed in Section \ref{intro}, the weakly-supervised activity localization and classification problem as addressed in this paper can be directly mapped to the problem of Multiple Instance Learning (MIL) \cite{zhou2004multi}. In MIL, individual samples are grouped in two bags, namely positive and negative bags. A positive bag contains at least one positive instance and a negative bag contains no positive instance. Using these bags as training data, we need to learn a model, which will be able to distinguish each instance to be positive or negative, besides classifying a bag. In our case, we consider the entire video as a bag of instances, where each instance is represented by a feature vector at a certain time instant. In order to compute the loss for each bag, i.e., video in our case, we need to represent each video using a single confidence score per category. For a given video, we compute the activation score corresponding to a particular category as the average of $k$-max activation over the temporal dimension for that category. As in our case, the number of elements in a bag varies widely, we set $k$ proportional to the number of elements in a bag. Specifically,
\begin{equation}
k_i=\max\Big(1,\left \lfloor{\dfrac{l_i}{s}}\right \rfloor \Big)
\label{k}
\end{equation}
where $s$ is a design parameter. Thus, our class-wise confidence scores for the $j^{th}$ category of the $i^{th}$ video can be represented as,
\begin{equation}
s_i^j = \frac{1}{k_i} \max_{
	\substack{\mathcal{M} \subset \boldsymbol{\mathcal{A}}_i[j,:] \\
		|\mathcal{M}| = k_i}}	
\sum_{l=1}^{k_i} \mathcal{M}_l
\label{scores}
\end{equation}
where $\mathcal{M}_l$ indicates the $l^{th}$ element in the set $\mathcal{M}$. Thereafter, a softmax non-linearity is applied to obtain the probability mass function over the all the categories as follows, $p^j_i=\frac{\exp(s^j_i)}{\sum_{j=1}^{n_c}\exp(s^j_i)}$. We need to compare this pmf with the ground truth distribution of labels for each video in order to compute the MILL. As each video can have multiple activities occurring in it, we represent the label vector for a video with ones at the positions if that activity occurs in the video, else zero. We then normalize this ground truth vector in order to convert it to a legitimate pmf. The MILL is then the cross-entropy between the predicted pmf $\boldsymbol{p}_i$ and ground-truth, which can then be represented as follows,
\begin{equation}
\mathcal{L}_{MILL} = \frac{1}{n}\sum_{i=1}^n \sum_{j=1}^{n_c} -y_i^j\log(p_i^j)
\end{equation}
where $\boldsymbol{y}_i=[y_i^1,\dots,y_i^{n_c}]^T$ is the normalized ground truth vector. This loss function is semantically similar to that used in \cite{wang2017untrimmednets}. We next present the novel Co-Activity Similarity Loss, which enforces constraints to learn better weights for activity localization.

\subsection{Co-Activity Similarity}
As discussed previously, the W-TALC problem motivates us to identify the correlations between videos of similar categories. Before discussing in more detail, let us define category-specific sets for the $j^{th}$ category as, $\mathcal{S}_j=\{\boldsymbol{x}_i \ | \exists \ a_i^k \in \boldsymbol{a}_i, \text{s.t.} \ a_i^k= \alpha_j  \}$, 
i.e., the set $\mathcal{S}_j$ contains all the videos of the training set, which has activity $\alpha_j$ as one of its labels. Ideally, we may want the following properties in the learned feature representations $\boldsymbol{X}_i$ in Eqn. \ref{featrep}.

\begin{itemize}
	\item A video pair belonging to the set $\mathcal{S}_j$ (for any $j \in \{1, \dots, n_c\}$) should have similar feature representations in the portions of the video where the activity $\alpha_j$ occurs. 
	\item For the same video pair, feature representation of the portion where $\alpha_j$ occurs in one video should be different from that of the other video where $\alpha_j$ does not occur.
\end{itemize} 
These properties are not directly enforced in MILL. Thus, we introduce Co-Activity Similarity Loss to embed the desired properties in the learned feature representations. As we do not have frame-wise labels, we use the class-wise activations obtained in Eqn. \ref{act} to identify the required activity portions. The loss function is designed in a way which helps to learn simultaneously the feature representation as well as the label space projection. We first normalize the per-video class-wise activations scores along the temporal axis using softmax non-linearity as follows:
\begin{equation}
\boldsymbol{\hat{\mathcal{A}}}_i[j,t] = \frac{\exp(\boldsymbol{\mathcal{A}}_i[j,t])}{\sum_{t'=1}^{l_i}\exp(\boldsymbol{\mathcal{A}}_i[j,t'])}
\label{attn}
\end{equation}
where $t$ indicates the time instants and $j \in \{1, \dots, n_c\}$. We refer to these as \textit{attention}, as they attend to the portions of the video where an activity of a certain category occurs. A high value of attention for a particular category indicates its high occurrence-probability of that category. In order to formulate the loss function, let us first define the class-wise feature vectors of regions with high and low attention as follows: 
\begin{align}
^H\boldsymbol{\boldsymbol{f}}_i^j &= \boldsymbol{X}_i \boldsymbol{\hat{\mathcal{A}}}_i[j,:]^T \nonumber \\
^L\boldsymbol{\boldsymbol{f}}_i^j &= \frac{1}{l_i-1} \boldsymbol{X}_i \Big(\boldsymbol{1}-\boldsymbol{\hat{\mathcal{A}}}_i[j,:]^T\Big)
\label{highfeat} 
\end{align}
where $^H\boldsymbol{\boldsymbol{f}}_i^j, ^L\boldsymbol{\boldsymbol{f}}_i^j \in \mathbb{R}^{2048}$ represents the high and low attention region aggregated feature representations respectively of video $i$ for category $j$. It may be noted that in Eqn. \ref{highfeat} the low attention feature is not defined if a video contains a certain activity and the number of feature vectors, i.e., $l_i=1$. This is also conceptually valid and in such cases, we cannot compute the CASL. We use cosine similarity in order to obtain a measure of the degree of similarity between two feature vectors and it may be expressed as follows:
\begin{equation}
d[\boldsymbol{f}_i,\boldsymbol{f}_j] = 1 - \frac{\langle \boldsymbol{f}_i,\boldsymbol{f}_j \rangle}{\langle\boldsymbol{f}_i,\boldsymbol{f}_i\rangle^{\frac{1}{2}}\langle\boldsymbol{f}_j,\boldsymbol{f}_j\rangle^{\frac{1}{2}}}
\label{cosine}
\end{equation}
In order to enforce the two properties discussed above, we use the ranking hinge loss. Given a pair of videos $\boldsymbol{x}_m, \boldsymbol{x}_n \in \mathcal{S}_j$, the loss function may be represented as follows:
\begin{align}
\mathcal{L}^{mn}_j &= \frac{1}{2}\Big\{\max\Big(0, d[ ^H\boldsymbol{\boldsymbol{f}}_m^j, ^H\boldsymbol{\boldsymbol{f}}_n^j] - d[ ^H\boldsymbol{\boldsymbol{f}}_m^j, ^L\boldsymbol{\boldsymbol{f}}_n^j] + \delta\Big) \nonumber \\
&+ \max\Big(0, d\big[ {}^H \boldsymbol{\boldsymbol{f}}_m^j, {}^H \boldsymbol{\boldsymbol{f}}_n^j\big] - d\big[  {}^L\boldsymbol{\boldsymbol{f}}_m^j, ^H\boldsymbol{\boldsymbol{f}}_n^j\big] + \delta\Big)\Big\}
\label{pairloss}
\end{align}
where $\delta$ is the margin parameter and we set it to $0.5$ in our experiments. The two terms in the loss function are equivalent in meaning, and they represent that the high attention region features in both the videos should be more similar than the high attention region feature in one video and the low attention region feature in the other video. The total loss for the entire training set may be represented as follows:
\begin{equation}
\mathcal{L}_{CASL} = \frac{1}{n_c}\sum_{j=1}^{n_c} \frac{1}{{|\mathcal{S}_j| \choose 2}} \sum_{\boldsymbol{x}_m, \boldsymbol{x}_n \in \mathcal{S}_j} \mathcal{L}^{mn}_j
\label{col}
\end{equation}

\textbf{Optimization.} The total loss function we need to optimize in order to learn the weights of the weakly supervised layer can be represented as follows:
\begin{equation}
\mathcal{L} = \lambda\mathcal{L}_{MILL} + (1-\lambda)\mathcal{L}_{CASL} + \alpha  ||\boldsymbol{W}||_F^2
\label{loss}
\end{equation}
where the weights to be learned in our network are lumped to $\boldsymbol{W}$. We use $\lambda=0.5$ and $\alpha=5 \times 10^{-4}$ in our experiments. We optimize the above loss function using Adam \cite{kingma2014adam} with a batch size of 10. We create each batch in a way such that it has a minimum of three pairs of videos such that each pair has at least one category in common. We use a constant learning rate of $10^{-4}$ in all our experiments.

\textbf{Classification and Localization.} After learning the weights of the network, we use them to classify an untrimmed video as well as localize the activities in it during test time. Given a video, we obtain the class-wise confidence scores as in Eqn. \ref{scores} followed by softmax to obtain a pmf over the possible categories. Then, we can threshold the pmf to classify the video to contain one or more activity categories. However, as defined by the dataset \cite{idrees2017thumos} and used in literature \cite{wang2017untrimmednets}, we use mAP for comparison, which does not require the thresholding operation, but directly uses the pmf. 

For localization, we employ a two-stage thresholding scheme. First, we discard the categories which have confidence score (Eqn. \ref{scores}) below a certain threshold (0.0 used in our experiments). Thereafter, for each of the remaining categories, we apply a threshold on the corresponding activation in $\boldsymbol{\mathcal{A}}$ (Eqn. \ref{act}) along the temporal axis to obtain the localizations. It may be noted that as $l_i$ is generally less than the frame rate of the videos, we upsample the activations to meet the frame rate. 

\section{Experiments}
\label{exp}
In this section, we experimentally evaluate the proposed framework for activity localization and classification from weakly labeled videos. We first discuss the datasets we use, followed by the implementation details, quantitative and some qualitative results.

\textbf{Datasets.} 
We perform experimental analysis on two datasets namely ActivityNet v1.2 \cite{heilbron2015activitynet} and Thumos14 \cite{idrees2017thumos}. These two datasets contain untrimmed videos with frame-wise labels of activities occurring in the video. However, as our algorithm is weakly-supervised, we use only the activity tags associated with the videos.

\textit{ActivityNet1.2.} This dataset has $4819$ videos for training, $2383$ videos for validation and $2480$ videos for testing whose labels are withheld. The number of classes involved is $100$, with an average of $1.5$ temporal activity segments per video. As in literature \cite{wang2017untrimmednets,nguyen2017weakly}, we use the training videos to train our network, and the validation set to test. 

\textit{Thumos14.} The Thumos14 dataset has $1010$ validation videos and $1574$ test videos divided into $101$ categories. Among these videos, $200$ validation videos and $213$ test videos have temporal annotations belonging to $20$ categories. Although this is a smaller dataset than ActivityNet1.2, the temporal labels are very precise and with an average of 15.5 activity temporal segments per video. This dataset has several videos where multiple activities occur, thus making it even more challenging. The length of the videos also varies widely from a few seconds to more than an hour. The lesser number of videos make it challenging to efficiently learn the weakly-supervised network. Following literature \cite{wang2017untrimmednets,nguyen2017weakly}, we use the validation videos for training and the test videos for testing.

\textbf{Implementation Details.} We use the corresponding repositories to extract the features for UntrimmedNets\footnote{www.github.com/wanglimin/UntrimmedNet} and I3D\footnote{www.github.com/deepmind/kinetics-i3d}. We do not finetune the feature extractors. The weights of the weakly supervised layers are initialized by Xavier method \cite{glorot2010understanding}. We use TVL1 optical flow  \footnote{www.github.com/yjxiong/temporal-segment-networks}. We train our network on a single Tesla K80 GPU using Tensorflow. We set $s=8$ in Eqn. \ref{k} for both the datasets.

\begin{table}[h!]
	\fontsize{8.5}{9.5}\selectfont
	\centering	
	\caption{Detection performance comparisons over the Thumos14 dataset. UNTF and I3DF are abbreviations for UntrimmedNet features and I3D features respectively. The symbol $^\downarrow$ represents that following \cite{nguyen2017weakly}, those models are trained using only the $20$ classes having temporal annotations, but without using their temporal annotations.}
	\begin{tabular}{C{1.9cm} || p{2.8cm} || C{1cm} C{1cm}  C{1cm}  C{1cm}  C{1cm} C{1cm}}
		\hline
		{Supervision} & {IoU $\rightarrow$} & {0.1} & {0.2} & {0.3} & {0.4} & {0.5} & {0.7}\\
		\hline \hline
		{\multirow{10}{2.5em}{Strong}} & {Saliency-Pool \cite{karaman2014fast}} & {04.6} & {03.4} & {02.1} & {01.4} & {00.9} & {00.1}\\
		& {FV-DTF \cite{oneata2014lear}} & {36.6} & {33.6} & {27.0} & {20.8} & {14.4} & {-}\\
		& {SLM-mgram \cite{richard2016temporal}} & {39.7} & {35.7} & {30.0} & {23.2} & {15.2} & {-}\\
		& {S-CNN \cite{shou2016temporal}} & {47.7} & {43.5} & {36.3} & {28.7} & {19.0} & {05.3}\\
		& {Glimpse \cite{yeung2016end}} & {48.9} & {44.0} & {27.0} & {20.8} & {14.4} & {-}\\
		& {PSDF \cite{yuan2016temporal}} & {51.4} & {42.6} & {33.6} & {26.1} & {18.8} & {-}\\
		& {SMS \cite{yuan2017temporal}} & {51.0} & {45.2} & {36.5} & {27.8} & {17.8} & {-}\\
		& {CDC \cite{shou2017cdc}} & {-} & {-} & {40.1} & {29.4} & {23.3} & {\textbf{07.9}}\\
		& {R-C3D \cite{xu2017r}} & {54.5} & {51.5} & {44.8} & {35.6} & {28.9} & {-}\\	
		& {SSN \cite{zhao2017temporal}} & {\textbf{60.3}} & {\textbf{56.2}} & {\textbf{50.6}} & {\textbf{40.8}} & {\textbf{29.1}} & {-}\\	
		\hline
		\hline
		{\multirow{4}{2.5em}{Weak}} & {HAS \cite{singh2017hide}} & {36.4} & {27.8} & {19.5} & {12.7} & {06.8} & {-} \\
		& {UntrimmedNets \cite{wang2017untrimmednets}} & {44.4} & {37.7} & {28.2} & {21.1} & {13.7} & {-}\\
		& {STPN (UNTF) \cite{nguyen2017weakly} $^\downarrow$} & {45.3} & {38.8} & {31.1} & {23.5} & {16.2} & {05.1}\\
		& {STPN (I3DF) \cite{nguyen2017weakly} $^\downarrow$} & {52.0} & {44.7} & {35.5} & {25.8} & {16.9} & {04.3}\\
		\hline
		{\multirow{4}{2.5em}{Weak (Ours)}} & {MILL+CASL+UNTF$^\downarrow$} & {\textbf{49.0}} & {\textbf{42.8}} & {\textbf{32.0}} & {\textbf{26.0}} & {\textbf{18.8}} & {\textbf{06.2}} \\
		& {MILL+I3DF} & {46.5} & {39.9} & {31.2} & {24.0} & {16.9} & {04.4}\\		
		& {MILL+CASL+I3DF} & {53.7} & {48.5} & {39.2} & {29.9} & {22.0} & {07.3} \\
		& {MILL+CASL+I3DF$^\downarrow$} & {\textbf{55.2}} & {\textbf{49.6}} & {\textbf{40.1}} & {\textbf{31.1}} & {\textbf{22.8}} & {\textbf{07.6}} \\
		\hline
		\hline
	\end{tabular}
	\label{thumos14_localization}
\end{table}
\begin{table}[h!]
	\fontsize{8.5}{9.5}\selectfont
	\centering	
	\caption{Detection performance comparisons over the ActivityNet1.2 dataset. The last column (Avg.) indicates the average mAP for IoU thresholds 0.5:0.05:0.95. }
	\begin{tabular}{C{1.6cm} || p{2.4cm} || C{1cm} C{1cm}  C{1cm}  C{1cm}  C{1cm} C{1cm} C{1cm}}
		\hline
		{Supervision} & {IoU $\rightarrow$} & {0.1} & {0.2} & {0.3} & {0.4} & {0.5} & {0.7} & {Avg.}\\
		\hline \hline	 
		{\multirow{2}{2.5em}{Strong}} & {SSN-SW \cite{zhao2017temporal}} & {-} & {-} & {-} & {-} & {-} & {-} & {24.8}\\
		& {SSN-TAG \cite{zhao2017temporal}} & {-} & {-} & {-} & {-} & {-} & {-} & {\textbf{25.9}}\\	
		\hline
		\hline
		{Weak} & {W-TALC (Ours)} & {\textbf{53.9}} & {\textbf{49.8}} & {\textbf{45.5}} & {\textbf{41.6}} & {\textbf{37.0}} & {\textbf{14.6}} & {\textbf{18.0}}\\
		\hline
		\hline
	\end{tabular}
	\label{activitynet_localization}
\end{table}

\textbf{Activity Localization.} We first perform a quantitative analysis of our framework for the task of activity localization. We use mAP with different Intersection over Union (IoU) thresholds as a performance metric, as in \cite{idrees2017thumos}. We compare our results with several state-of-the-art methods on both strong and weak supervision in Table \ref{thumos14_localization} and \ref{activitynet_localization} for Thumos14 and ActivityNet1.2 respectively. It may be noted that to the best of our knowledge, we are first to present quantitative results on weakly-supervised temporal activity localization on ActivityNet1.2. We show results for different combinations of features and loss function used. It may be noted that our framework performs much better than the other weakly supervised methods with similar feature usage. It is important to note that although the Kinetics pre-trained I3D features (I3DF) have some knowledge about activities, using only MILL as in \cite{wang2017untrimmednets} along with I3DF performs much worse than combining it with CASL, which is introduced in this paper. Moreover, our framework performs much better than other state-of-the-art methods even when using UNTF, which is not trained using any strong labels of activities. A detailed analysis of the two loss functions MILL and CASL will be presented subsequently.  

\textbf{Activity Classification.}
We now present the performance of our framework for activity classification. We use mean average precision (mAP) to compute the classification performance from the predicted videos-level scores in Eqn. \ref{scores} after applying softmax. We compare with both fully supervised and weakly-supervised methods and the results are presented in Table \ref{thumos_classification} and \ref{activitynet_classification} for Thumos14 and ActivityNet1.2 respectively. The proposed method performs significantly better than the other state-of-the-art approaches. Please note that the methods indicated with $^\uparrow$ utilize a larger training set compared to ours as mentioned in the tables. 

\begin{table}[t!]
	\begin{minipage}[h]{0.48\textwidth}
		\caption{Classification performance comparisons over Thumos14 dataset. $^\uparrow$ indicates that the algorithm use both videos from Thumos14 and trimmed videos from UCF101 for training. Without $^\uparrow$ indicates that the algorithm uses only videos from Thumos14 for training.}
		\fontsize{8.5}{9.5}\selectfont
		\begin{tabular}{ l || c || c } 
			\hline
			{Methods} & {mAP} & {Supervision} \\
			\hline \hline
			{EMV + RGB \cite{zhang2016real}} & {61.5} & {Strong $^\uparrow$}\\
			{iDT+FV \cite{wang2013action}} & {63.1} & {Strong $^\uparrow$}\\
			{iDT+CNN \cite{wang2014action}} & {62.0} & {Strong $^\uparrow$}\\
			{Objects + Motion \cite{jain201515}} & {71.6} & {Strong $^\uparrow$}\\
			{Feat. Agg. \cite{jain2014university}} & {71.0} & {Strong $^\uparrow$}\\
			{Extreme LM \cite{varol2015efficient}} & {63.2} & {Strong $^\uparrow$}\\
			{Temp. Seg. Net. (TSN) \cite{wang2016temporal}} & {78.5} & {Strong $^\uparrow$}\\
			{Two Stream  \cite{simonyan2014two}} & {66.1} & {Strong $^\uparrow$}\\
			{Temp. Seg. Net. (TSN) \cite{wang2016temporal}} & {67.7} & {Strong}\\		
			\hline
			\hline
			{UntrimmedNets \cite{wang2017untrimmednets}} & {74.2} & {Weak}\\
			{UntrimmedNets \cite{wang2017untrimmednets}} & {82.2} & {Weak $^\uparrow$}\\
			{W-TALC (Ours w. I3D)} & {\textbf{85.6}} & {Weak}\\
			\hline
			\hline
		\end{tabular}
		\label{thumos_classification}
	\end{minipage}
	\hfill
	\begin{minipage}{0.48\textwidth}
		\caption{Classification performance comparisons over the ActivityNet1.2 dataset. $^\uparrow$ indicate that the algorithm use the training and validation set of ActivityNet1.2 for training and tested on the server. Without $^\uparrow$ means that the algorithm is trained on the training set and tested on the validation set.}
		\begin{tabular}{l | c | c} 
			\hline
			{Algorithms} & {mAP} & {Supervision} \\
			\hline \hline
			{C3D \cite{tran2015learning}} & {74.1} & {Strong $^\uparrow$}\\
			{iDT+FV \cite{wang2013action}} & {66.5} & {Strong $^\uparrow$}\\
			{Depth2Action \cite{jain201515}} & {78.1} & {Strong $^\uparrow$}\\
			{Temp. Seg. Net. (TSN) \cite{wang2016temporal}} & {88.8} & {Strong $^\uparrow$}\\
			{Two Stream  \cite{simonyan2014two}} & {71.9} & {Strong $^\uparrow$}\\
			{Temp. Seg. Net. (TSN) \cite{wang2016temporal}} & {86.3} & {Strong}\\		
			\hline
			\hline
			{UntrimmedNets \cite{wang2017untrimmednets}} & {87.7} & {Weak}\\
			{UntrimmedNets \cite{wang2017untrimmednets}} & {91.3} & {Weak $^\uparrow$}\\
			{W-TALC (Ours w. I3D)} & {\textbf{93.2}} & {Weak}\\
			\hline
			\hline
		\end{tabular}
		\label{activitynet_classification}
	\end{minipage}
\end{table}

\begin{figure*}
	\centering
	\begin{subfigure}{0.48\textwidth}
		\includegraphics[scale=0.4]{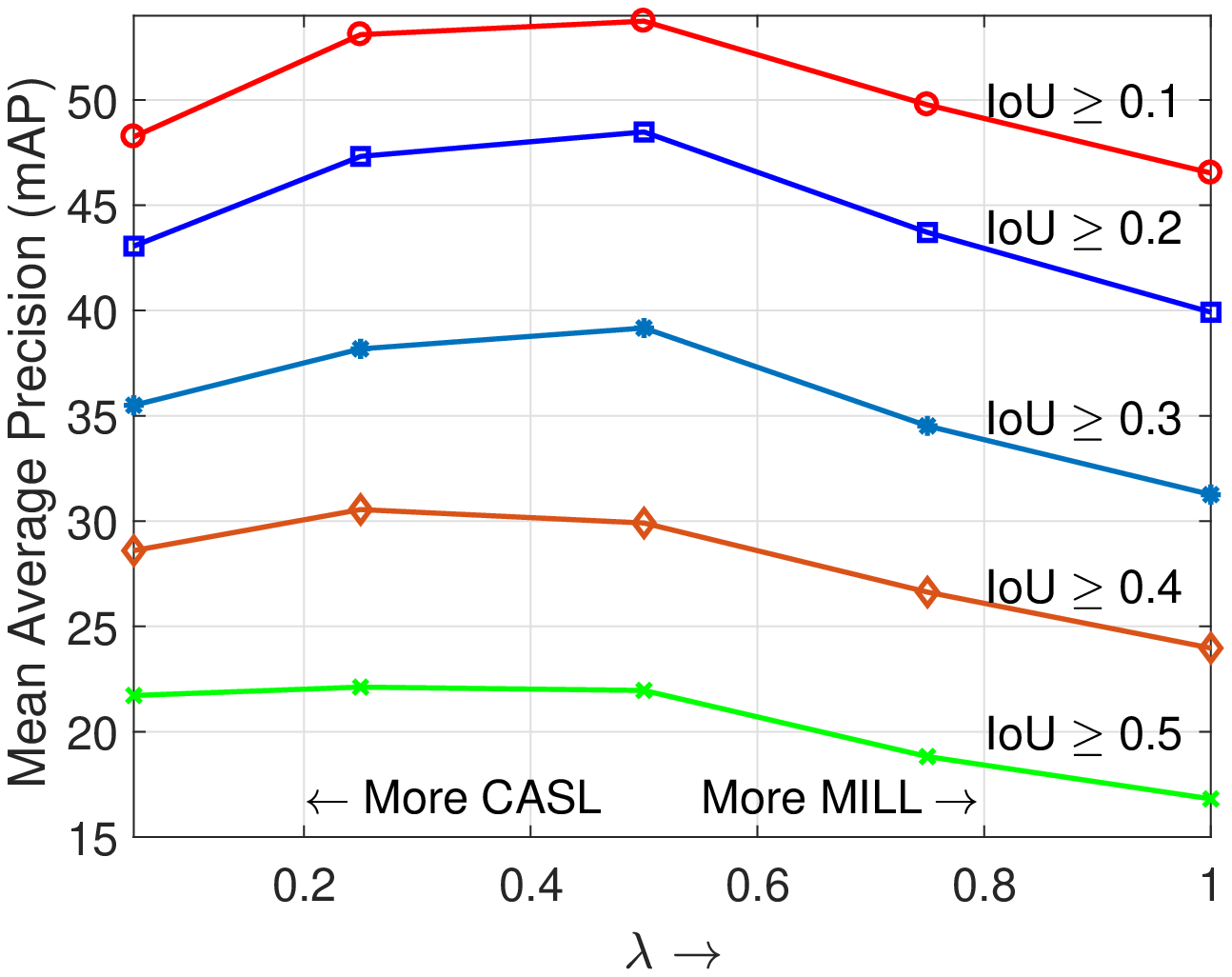}
		\caption{}
		\label{lambda}
	\end{subfigure}
	\begin{subfigure}{0.48\textwidth}
		\includegraphics[scale=0.4]{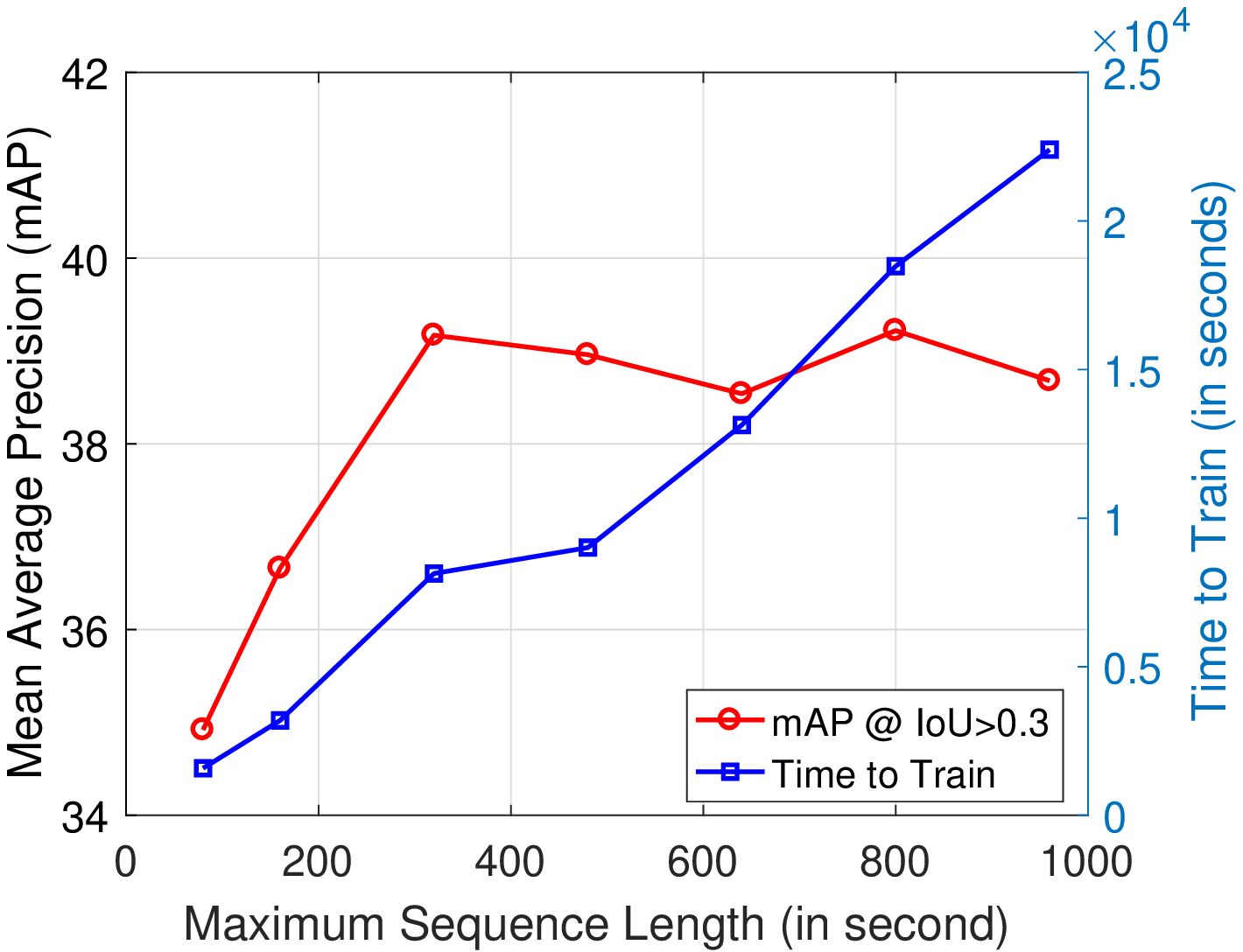}
		\caption{}
		\label{time}
	\end{subfigure}
	\caption{(a) presents the variations in detection performance on Thumos14 by changing weights on MILL and CASL. Higher $\lambda$ represents more weight on the MILL and vice versa. (b) presents the variations in detection performance ($@ \text{IoU} \geq 0.3$) and training time on Thumos14 dataset by changing the maximum possible length of video sequence during training ($T$) as discussed in the text.}
	\label{expanalysis}
\end{figure*}

\textbf{Relative Weights on Loss Functions.} In our framework, we jointly optimize two loss functions - MILL and CASL defined in Eqn. \ref{loss} to learn the weights of the weakly-supervised module. It is interesting to investigate the relative contributions of the loss functions to the detection performance. In order to do that, we performed experiments, using the I3D features, with different values of $\lambda$ (higher value indicate larger weight on MILL) and present the detection results on the Thumos14 dataset in Fig. \ref{lambda}. 

As may be observed from the plot, the proposed method performs best with $\lambda=0.5$, i.e., when both the loss functions have equal weights. Moreover, using only MILL, i.e., $\lambda=1.0$, results in a decrease of $7-8\%$ in mAP compared to when both CASL and MILL are given equal weights in the loss function. This shows that the CASL introduced in this work has a major effect towards the better performance of our framework compared to using I3D features along with the loss function in \cite{wang2017untrimmednets}, i.e., MILL.

\textbf{Sensitivity to Maximum Length of Sequence.}
Natural videos may often be very long. As mentioned previously, in the weakly-supervised setting, we have only video-level labels, so we need to process the entire video at once in order to compute the loss functions. In Section \ref{feature}, we discuss a simple sampling strategy, which we use to maintain the length of the videos in a batch to be less than a pre-defined length $T$ to meet GPU memory constraints. This method has the following advantages and disadvantage. 

\noindent
- \textit{Advantages}: First, we can learn from long length videos using this scheme. Secondly, this strategy will act as a data augmentation technique as we randomly crop, along the temporal axis to make it a fixed length sequence, if the length of the video $\geq T$. Also a lower value of $T$ reduces computation time.

\noindent
- \textit{Disadvantage}: In this sampling scheme, errors will be introduced in the labels of the training batch, which may increase with the number of training videos with length $> T$. \\
The above factors induce a trade-off between performance and computation time. This can be seen in Figure \ref{time}, wherein the initial portion of the plot, with an increase of $T$, the detection performance improves, but the computational time increases. However, the detection performance eventually reaches a plateau suggesting $T=320 s$ to be a reasonable choice for this dataset.

\begin{figure}[!t]
	\centering
	\includegraphics[scale=0.35]{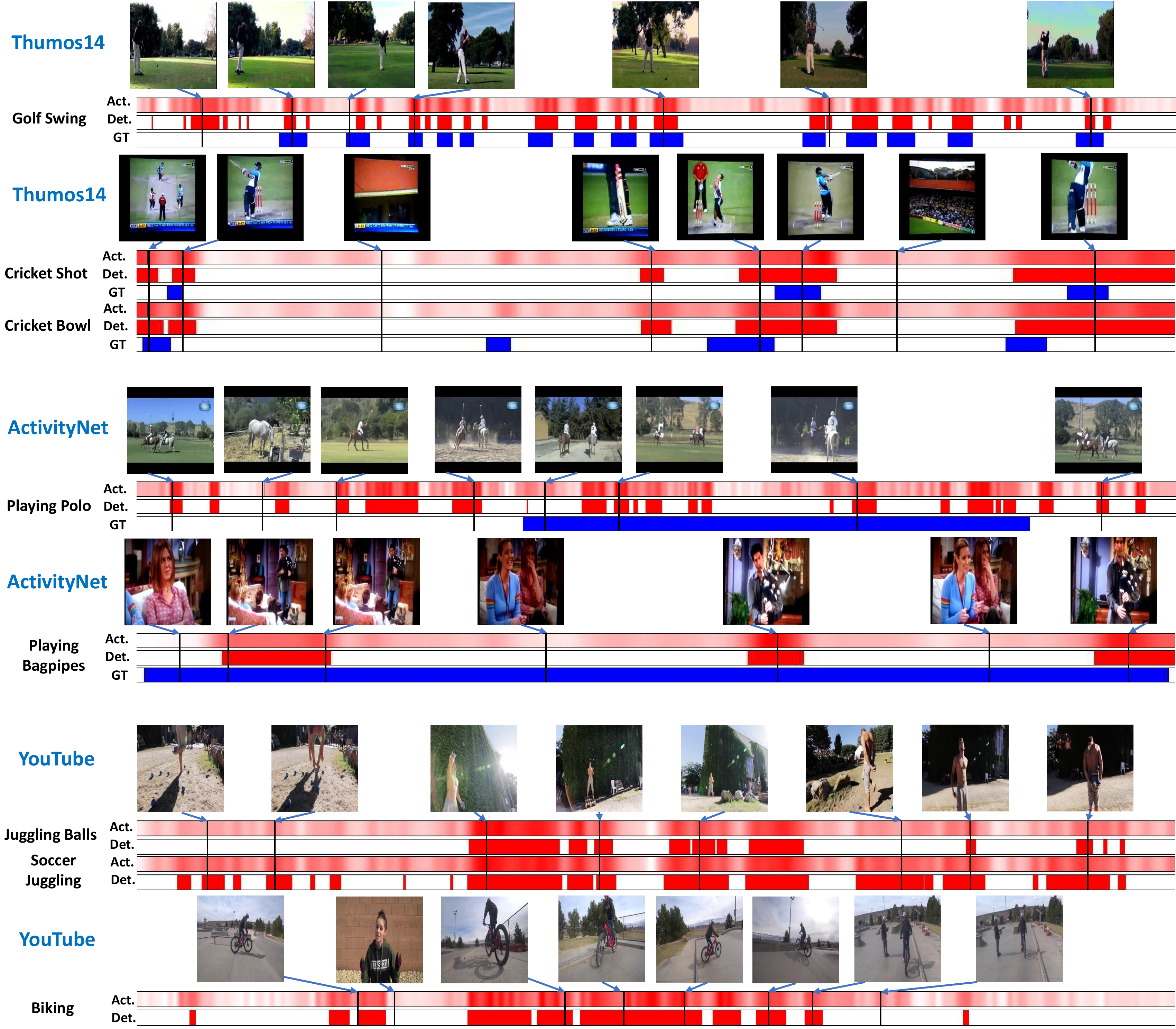}
	\caption{This figure presents some detection results for qualitative analysis. `Act.' represents the temporal activations obtained from the final layer of our network, `Det.' represents the detections obtained after thresholding the activations, and `GT' represent the ground truth. }
	\label{vis}
\end{figure}

\textbf{Qualitative Results.} We present a few interesting example localizations with ground truths in Fig. \ref{vis}. The figure has four examples from Thumos14 and ActivityNet1.2 datasets. To test how the proposed framework performs on videos outside the datasets used in this paper, we tested the learned networks on randomly collected videos from YouTube. We present two such example detections in Fig. \ref{vis}, using the model trained on Thumos14. 

The first example in Fig. \ref{vis} is quite challenging as the localization should precisely be the portions of the video, where Golf Swing occurs, which has very similar features in the RGB domain to portions of the video where the player prepares for the swing. In spite of this, our model is able to localize the relevant portions of Golf Swing, potentially based on the flow features. In the second example from Thumos14, the detections of Cricket Shot and Cricket Bowl appear to be correlated in time. This is because Cricket Shot and Bowl are two activities which generally co-occur in videos. To have fine-grained localizations for such activities, videos which have only one of these activities are required. However, in the Thumos14 dataset, very few training examples contain only one of these two activities. 

In the third example, which is from ActivityNet1.2, although `Playing Polo' occurs in the first portion of the video, it is absent in the ground truth. However, our model is able to localize those activity segments as well. The same discussion is also applicable to the fourth example, where `Bagpiping' occurs in the frames in a sparse manner, and our model's response is aligned with its occurrence, but the ground truth annotations are for almost the entire video. These two examples are motivations behind weakly-supervised localization, because obtaining precise unanimous ground truths from multiple labelers is difficult, costly and sometimes even infeasible.

The fifth example is on a randomly selected video from YouTube. It has a person, who is juggling balls in an outdoor environment. But, most of the examples in Thumos14 of the same category are indoors, with the person taking up a significant portion of the frames spatially. Despite such differences in data, our model is able to localize some portions of the activity. However, the model also predicts some portions of the video to be `Soccer Juggling', which may be because its training samples in Thumos14 contains a combination of feet, hand, and head, and a subset of such movements are present in `Juggling Balls'. Moreover, it is interesting to note that the first two frames show some maneuver of a ball with feet and it is detected as `Soccer Juggling' as well.

\section{Conclusions and Future Work}
In this paper, we present an approach to learn temporal activity localization and video classification models using only weak supervision with video-level labels. We present the novel Co-Activity Similarity loss, which is empirically shown to be complimentary with the Multiple Instance Learning Loss. We also show a simple mechanism to deal with long length videos, yet processing them at high granularity. Experiments on two challenging datasets demonstrate that the proposed method achieves state-of-the-art results in the weak TALC problem. Future work will concentrate on extending the idea of Co-Activity Similarity Loss to other problems in computer vision.

\noindent
\textbf{Acknowledgments.} This work was partially supported by ONR contract N00014-15-C-5113 through a sub-contract from Mayachitra Inc and NSF grant IIS-1724341. We thank Victor Hill of UCR CS for setting up the computing infrastructure.

\bibliographystyle{splncs04}
\bibliography{egbib}

\end{document}